\documentclass[10pt,twocolumn,letterpaper]{article}

\usepackage{cvpr}
\usepackage{times}
\usepackage{epsfig}
\usepackage{graphicx}
\usepackage{amsmath}
\usepackage{amssymb}

\usepackage[ruled, vlined]{algorithm2e} 
\usepackage{bm}
\usepackage{enumitem}
\usepackage{makecell}
\usepackage{multirow}
\usepackage{tabularx}
\usepackage{xcolor}
\usepackage[normalem]{ulem}

\DeclareMathOperator*{\argmax}{arg\,max}

\def\showremoved{1}
\newcommand{\letsremove}[1]{\if\showremoved1{\textcolor{blue}{\sout{#1}}}\fi}

\usepackage[pagebackref=true,breaklinks=true,letterpaper=true,colorlinks,bookmarks=false]{hyperref}

\cvprfinalcopy 

\def\cvprPaperID{1130} 

\ifcvprfinal\fi
\begin{document}

\title{Linguistic Structures as Weak Supervision for Visual Scene Graph Generation}

\author{Keren Ye \hspace{1cm} Adriana Kovashka\\
Department of Computer Science, University of Pittsburgh\\
{\tt\small \{yekeren, kovashka\}@cs.pitt.edu
}\\
{\small \url{https://github.com/yekeren/WSSGG}
}
}

\maketitle
\thispagestyle{empty}

\begin{abstract}
Prior work in scene graph generation requires categorical supervision at the level of triplets---subjects and objects, and predicates that relate them, either with or without bounding box information. However, scene graph generation is a holistic task: thus holistic, contextual supervision should intuitively improve performance. In this work, we explore how linguistic structures in captions can benefit scene graph generation. Our method captures the information provided in captions about relations between individual triplets, and context for subjects and objects (e.g. visual properties are mentioned). Captions are a weaker type of supervision than triplets since the alignment between the exhaustive list of human-annotated subjects and objects in triplets, and the nouns in captions, is weak. However, given the large and diverse sources of multimodal data on the web (e.g. blog posts with images and captions), linguistic supervision is more scalable than crowdsourced triplets. We show extensive experimental comparisons against prior methods which leverage instance- and image-level supervision, and ablate our method to show the impact of leveraging phrasal and sequential context, and techniques to improve localization of subjects and objects.
\end{abstract}

\vspace{-0.3cm}
\section{Introduction}
\label{sec:intro}
\vspace{-0.1cm}

Recognizing visual entities and understanding the relations among them are two fundamental problems in computer vision. The former task is known as object detection (OD) and the latter as visual relation detection (VRD). In turn, scene graph generation (SGGen) requires to jointly detect the entities and classify their relations.

\begin{figure}[t]
    \centering
    \includegraphics[width=1\linewidth]{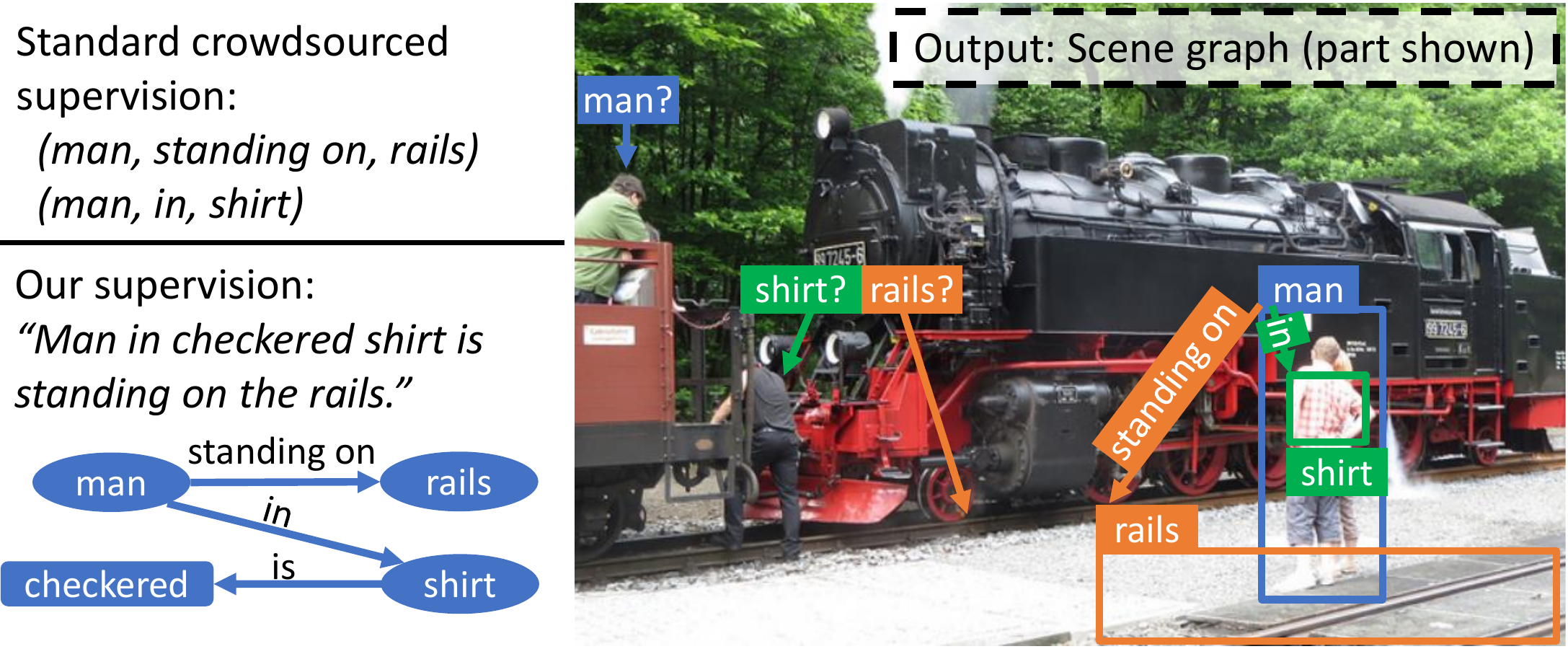}
    \caption{We tackle the problem of generating scene graphs with supervision in the form of captions at training time. 
    Parsing from captions enables utilization of the huge amount of image-text data available on the internet. The linguistic structure extracted maintains the relational information described in the caption without the loss of cross-triplet references, and facilitates disambiguation. 
    }
    \label{fig:concept}
\end{figure}

While scene graphs are a holistic, contextual representation of an image, the types of supervision that have been used capture context in an impoverished way. In particular, prior methods use supervision in the form of either subject-predicate-object triplets with bounding boxes for the subject and object \cite{lu2016visual,Newell_2017_NeurIPS,Yang_2018_ECCV} or subject-predicate-object triplets at the image level only \cite{Zareian_2020_CVPR,Zhang_2017_ICCV}. Thus, information in the supervision is local (separate triplets) while the scene graph to be output captures the entire image. This discrepancy between the properties of the desired output (global) and training data (local) becomes problematic due to potential ambiguity in the visual input. For example, in Fig.~\ref{fig:concept}, multiple \emph{persons} are standing on the \emph{rails}. Thus, standard supervision (top) which breaks down a scene graph into triplets, may create confusion.

In contrast, \emph{captions} capture global context that allows us to link multiple triplets, and localize a man who is both standing on the rails, and wearing a (checkered) shirt. 
Captions are linguistic constructs, and language could be argued to capture common sense (e.g., BERT \cite{devlin2019bert} models are good at question-answering and commonsense tasks).
Captions are also advantageous in terms of cost: humans naturally provide language descriptions of visual content they upload, thus caption-like supervision can be seen as ``free''.
However, caption supervision contains noise, which presents some challenges. First, captions provide supervision at the image level, similar to prior work in weakly-supervised scene graph generation \cite{Zareian_2020_CVPR}. Second, prior work \cite{misra2016seeing,Ye_2019_ICCV} shows that captions do not cover all relevant objects: not all content is mentioned, and some of the mentioned content is not referring to the image explicitly or is not easily localizable. 
Because captions are noisy, the supervision we use is even weaker than prior work \cite{Zareian_2020_CVPR}.

We propose an approach that leverages global context, using captions as supervision. Our approach models context for scene graphs in two ways. First, it extracts information from captions beyond the subject-predicate-object entities (e.g., in the form of attributes like ``checkered'', in Fig.~\ref{fig:concept}). This context enables more accurate representations of concepts, and thus more accurate localization of each subject-predicate-object triplet. Second, visuo-linguistic context provides a way to reason about common-sense relationships within each triplet, to prevent non-sensical triplets from being generated (e.g., ``rails standing on man'' is unlikely, while ``man standing on rails'' is likely). 
To cope with the challenges of the noise contained in captions, we rely on an iterative detection method which helps prune some spurious relations between caption words and image regions, via boostrapping.
While the captions we use are crowdsourced, our method paves the road for using image-caption pairs harvested from the internet for free, using text accompanying images on the web, from blogs, social media posts, YouTube video descriptions, and instructional videos \cite{miech2019howto100m,sharma-etal-2018-conceptual,young-etal-2014-image}. 
Note that our method internally uses a graph with broad types of nodes, including adjectives, even though these are not part of the graph that is being output at test time.
A side contribution is an adaptation of techniques from weakly-supervised object detection to improve localization of subject and object through iterative refinement, which has not been used for scene graph generation before.

To isolate the contribution of global context from the noise contained in captions (i.e., objects not being mentioned), we verify our approach in two settings. First, we construct a ground-truth triplet graph by connecting triplets with certain overlap. We show that our full method greatly outperforms prior work (it boosts the performance of \cite{Zareian_2020_CVPR} by 59\%-67\%). 
Second, we use two types of actual captions. This causes overall performance to drop, but we observe that modeling phrasal (cross-triplet) and sequential (within-triplet) linguistic context achieves strong results, significantly better than more direct uses of captions, and competitive with methods using clean image-level supervision.

To summarize, our contributions are as follows:
\begin{itemize}[nolistsep,noitemsep]
    \item We examine a new mechanism for scene graph generation using a new type of weak supervision.
    \item We contextualize embeddings for subject/object entities based on linguistic structures (e.g. noun phrases). 
    \item We propose new joint classification and localization of subject, object and predicate within a triplet.
    \item We leverage weakly-supervised object detection techniques to improve scene graph generation.
\end{itemize}

\section{Related Work}
\label{sec:related}
\vspace{-0.1cm}

\begin{figure*}[t]
    \centering
    \includegraphics[width=1.0\linewidth]{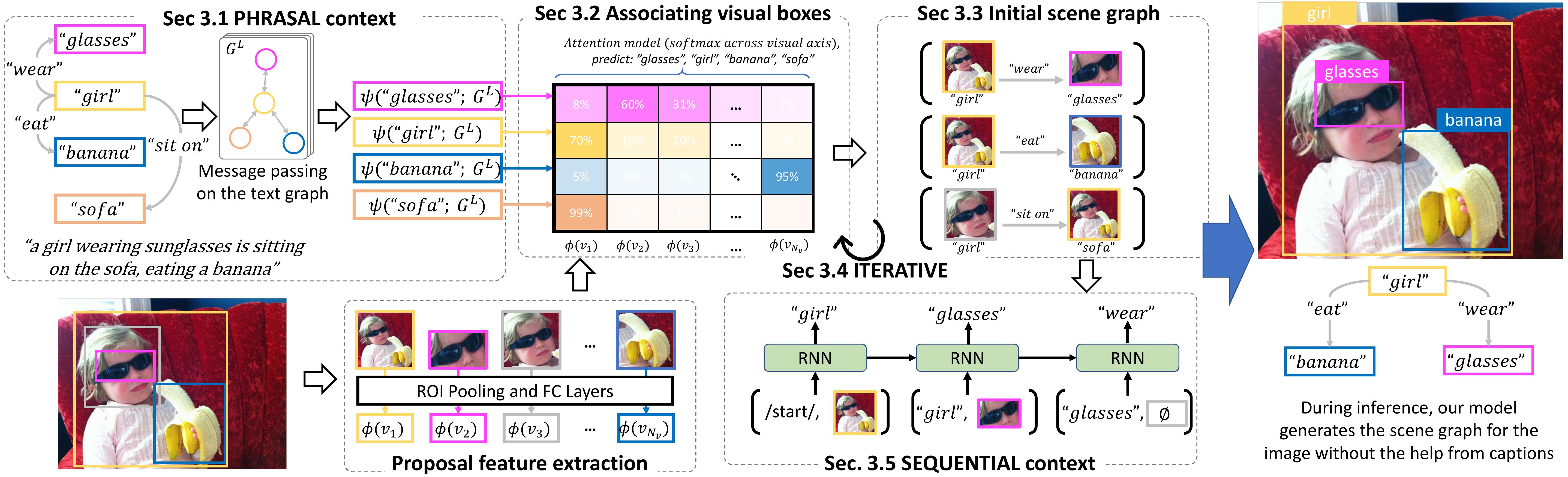}  
    \caption{Model overview.
    Our model uses the image's paired caption as weak supervision to learn the entities in the image and the relations among them. At inference time, it generates scene graphs without help from texts.
    To learn our model, we first allow context information to propagate on the text graph to enrich the entity word embeddings (Sec.~\ref{sec:approach:phrasal_context}). We found this enrichment provides better localization of the visual objects.
    Then, we optimize a text-query-guided attention model (Sec.~\ref{sec:approach:grounding}) to provide the image-level entity prediction and associate the text entities with visual regions best describing them.
    We use the joint probability (Eq.~\ref{eq:initial_sggen}) to choose boxes associated with both subject and object (Sec.~\ref{sec:approach:scene_graph}), then use the top scoring boxes (Eq.~\ref{eq:oicr}) to learn better grounding (Sec.~\ref{sec:approach:iterative}).
    Finally, we use an RNN (Sec.~\ref{sec:approach:sequential_context}) to capture the vision-language common-sense and refine our predictions.
    Our code is available at \url{https://github.com/yekeren/WSSGG}. 
    }
    \label{fig:overview}
\end{figure*}

\textbf{Learning from textual descriptions:} 
Open information extraction systems ~\cite{angeli-etal-2015-leveraging,del2013clausie,fader-etal-2011-identifying,mausam-etal-2012-open,yates-etal-2007-textrunner} produce relation triples using surface and dependency patterns, but target language-only relation extraction or question answering. On the vision end, method exist to parse a question or image into a structured, tree-like form, for composable visual reasoning \cite{Andreas_2016_CVPR,goldman-etal-2018-weakly,Johnson_2017_ICCV,Kim_2020_CVPR,Mao2019NeuroSymbolic,yi2018neural}.
Following the emergence of scene graphs~\cite{Johnson_2015_CVPR} as a global description of an image, automatic parsing from textual descriptions to scene graphs \cite{schuster-etal-2015-generating,wang-etal-2018-scene} aims to fill the gap between texts and images. It tackles practical issues such as pronoun resolution and plural nouns, and duplicates some nodes in the scene graph if necessary.
Though we use the parser designed in \cite{schuster-etal-2015-generating}, our reliance on parsing is different. While the above methods tackle pure language tasks, visual question answering, and image retrieval, we use the parsed results as supervised signals to guide a scene graph generation model during training. Our work is similar to \cite{Chen_2017_CVPR,jerbi2020learning,Ye_2019_ICCV} since we extract or amplify information from captions. However, these works only extract \emph{entities} from captions, while we also learn from the properties and relations described.
Also related are recent methods that use supervision from visual-language pairs  \cite{desai2020virtex,Miech_2020_CVPR,Nagrani_2020_CVPR,suris2020learning,Wu_2019_CVPR}, but these learn general-purpose representations and do not perform scene graph generation.

\textbf{Visual grounding of phrases}
locates the entities in an image, based on a given natural language query. \cite{karpathy2014deep} align sentence fragments with image regions. \cite{Chen_2018_CVPR, rohrbach2016grounding} attend to the relevant image regions to reconstruct the input phrase, similar to  weakly-supervised object detection. \cite{Zhao_2018_CVPR} incorporate a spatial transformer~\cite{jaderberg2015spatial} to refine object boxes relative to multi-scale anchors. We use a technique similar to visual grounding to find label-related regions, but our key innovation lies in our use of the linguistic structure. We allow context to propagate to language queries to improve entity detection. Our model only takes image inputs at test time.

\textbf{Scene graph generation (SGGen)} aims to localize and recognize all visual entities and predict predicates between them. Most approaches \cite{Chen_2019_CVPR,Gu_2019_CVPR,Li_2017_ICCV,Lin_2020_CVPR,lu2016visual,Newell_2017_NeurIPS,Qi_2019_CVPR,Xu_2017_CVPR,Yang_2018_ECCV,Zellers_2018_CVPR} learn to generate graphs in a fully-supervised manner, in which training data involves both entities (bounding boxes and labels) and predicates. Inspired by weakly-supervised object detection (WSOD)~\cite{Bilen_2016_CVPR,Oquab_2015_CVPR}, \cite{Peyre_2017_ICCV,Zareian_2020_CVPR,Zhang_2017_ICCV} somewhat reduce the reliance on these \emph{labor-intensive} annotations. \cite{Peyre_2017_ICCV} infer visual relations using only image-level triplets. \cite{Zhang_2017_ICCV} directly apply WSOD for entity localization and add a weakly-supervised visual relation detection (WSVRD) task for classifying entity pairs. \cite{Zareian_2020_CVPR} match predicates to entities and jointly infer the entities, predicates, and their alignments, using a bipartite graph. 
However, 
\cite{Peyre_2017_ICCV,Zareian_2020_CVPR,Zhang_2017_ICCV} still require clean triplet annotations from crowdsourcing, while our method only requires captions.
Further, we capture visual properties in the internal graph our method uses at training time; these cannot be represented using triplets but help to enrich the visual representation and better ground entities. 
\cite{Zareian_2020_CVPR}'s method includes a more general (subject, predicate, $\emptyset$) graph, but it does not capture visual attributes. 

\section{Approach}
\label{sec:approach}
\vspace{-0.1cm}


\textbf{Inputs.} Our method does not rely on dense human-annotated instances and relations, but takes in linguistic structures as supervised signals (Fig.~\ref{fig:overview} top-left). Such structural text information is abandoned in other weakly-supervised methods~\cite{Zareian_2020_CVPR,zhang2017visual,Zhang_2017_ICCV}. We first convert captions paired with images into text graphs using a language parser \cite{schuster-etal-2015-generating}. The resulting graphs describe the entities in the caption and the relations (e.g., verbs or prepositions) among them. 
We call this setting  \underline{Cap-Graph}.
Our method's performance depends on how exhaustive the caption is, and how robust is the parser chosen. Thus, we also design a setting where we extract a ground-truth text graph from the scene graph annotations, ignoring bounding boxes (\underline{GT-Graph}).

\begin{table*}[t]
    \scriptsize
    \centering
    \setlength\tabcolsep{1pt} 
    \begin{tabularx}{1\linewidth}{c | c}
        \Xhline{2\arrayrulewidth}
        {\begin{tabularx}{0.5\linewidth}{c}
            {\begin{tabularx}{0.5\linewidth}{p{2.6cm} | p{4.6cm} | p{1.4cm}}
                \multicolumn{3}{c}{Visual features} \\
            \Xhline{1\arrayrulewidth}
                $V_{prop}$ & Region proposals & $n_v \times 1$ \\
                $V_{feat}$ & Region proposal features & $n_v \times d_{cnn}$ \\
                $n_v=20$ & Number of region proposals \\
                $d_{cnn}=1536$ & Feature dimension \\
            \end{tabularx}} \\
            {\begin{tabularx}{0.5\linewidth}{p{2.6cm} | p{4.6cm} | p{1.4cm}}
            \Xhline{1\arrayrulewidth}
                \multicolumn{3}{c}{Text graph $G^L (E, R)$, parsed from caption} \\
            \Xhline{1\arrayrulewidth}
                $E=[e_i]_{i=1}^{n_e}$ & Entities (graph nodes) & $|E|=n_e$\\
                $R=[(r_i, s_i, o_i)]_{i=1}^{n_r}$ & Relations (graph edges) & $|R|=n_r$ \\
                $n_e, n_r$ & Number of entities/relations in a graph & \\
                $c_e, c_r$ & Number of entity/relation classes (vocab size) & \\
                $e_i$ & The $i$-th entity node, $e_i \in \{1 \cdots c_e\}$ & \\
                $r_i$ & The $i$-th relation edge, $r_i \in \{1 \cdots c_r\}$ & \\
                $s_i, o_i$ & Subject/object index of $i$-th relation, $s_i, o_i \in \{1 \cdots n_e\}$, $e_{s_i}, e_{o_i}$ refer to subject/object \\
            \end{tabularx}} \\
            {\begin{tabularx}{0.5\linewidth}{p{2.6cm} | p{4.6cm} | p{1.4cm}}
            \Xhline{1\arrayrulewidth}
                \multicolumn{3}{c}{Frozen GloVe embeddings} \\
            \Xhline{1\arrayrulewidth}
                $W_{ent}$ & Entity embedding matrix & $c_e \times d$ \\
                $W_{rel}$ & Relation embedding matrix & $c_r \times d$ \\
            \end{tabularx}}
        \end{tabularx}}
        &
        {\begin{tabularx}{0.5\linewidth}{c}
            {\begin{tabularx}{0.5\linewidth}{p{2.3cm} | p{5.5cm} | p{1.0cm}}
                \multicolumn{3}{c}{Image-level labels parsed from $G^L$} \\
            \Xhline{1\arrayrulewidth}
                $Y_{ent}$ & $Y_{ent}[i,:]$ is the one-hot representation of $e_i$ & $n_e \times c_e$  \\
                $Y_{rel}$ & $Y_{rel}[i,:]$ is the one-hot representation of $r_i$ & $n_r \times c_r$  \\
                $Y_{cssub}, Y_{csobj}$ & $Y_{cssub}[i,:], Y_{csobj}[i,:]$ are one-hot repr of $e_{s_i}, e_{o_i}$ & $n_r \times c_e$\\
                $Y_{cspred}$ & Alias of $Y_{rel}$ & $n_r \times c_r$\\
            \end{tabularx}} \\
            {\begin{tabularx}{0.5\linewidth}{p{2.3cm} | p{5.5cm} | p{1.0cm}}
            \Xhline{1\arrayrulewidth}
                \multicolumn{3}{c}{Instance-level pseudo labels} \\
            \Xhline{1\arrayrulewidth}
                $n_t$ & Number of iterations to improve $\bm{g}$ \\
                $\bm{g}^{(t)}, t\in \{0 \cdots n_t\}$ & Grounding vector, if $E$=$[girl, banana]$, $\bm{g}$=$[10, 17]$ means proposal $v_{10}$ is class \emph{girl} and $v_{17}$ is \emph{banana}  & $n_e \times 1$ \\
                $Y_{det}^{(t)},  t\in \{0 \cdots n_t\}$ & Entity detection label, $Y_{det}[i,j]$=$1$ means the proposal $v_i$ involves the $j$-th entity class & $n_v \times c_e$\\
                $Y_{relsub}, Y_{relobj}$ & Relation detection label, $Y_{relsub}[i,j]$=$1$ means the proposal $v_i$ may serve as a subject, and can apply the $j$-th relation to an unknown object; $Y_{relobj}[i,j]$=$1$ means the proposal $v_i$ may serve as an object, some unknown subject can apply the $j$-th relation to $v_i$ & $n_v \times c_r$\\
            \end{tabularx}}
        \end{tabularx}} \\
        \Xhline{2\arrayrulewidth}
    \end{tabularx}
    \caption{Overview of notation 
    for the visual features, linguistic structure $G^L$ and supervision parsed from $G^L$.}
    \label{tab:notations}
\end{table*}

\textbf{Training pipeline overview (Fig.~\ref{fig:overview}):} 
We extract the visual object proposals using FasterRCNN \cite{ren2015faster}.
We extract the text graph from paired captions (Cap-Graph) 
or directly read the ground-truth text graph (GT-Graph). We use a graph neural network based on the phrasal structure to enrich the text node representation (Fig.~\ref{fig:overview} top-left, Sec.~\ref{sec:approach:phrasal_context}). This enrichment simplifies the later localization step because we can search for more specifically described regions (e.g., ``girl eating banana,'' rather than ``girl'').
By optimizing the image-level entity scores and treating the text entities as queries, we obtain attention scores, which strongly imply the visual regions that best describe the text entities (Fig.~\ref{fig:overview} top-middle, Sec.~\ref{sec:approach:grounding}).
We design a way to learn from the weak signal of the attention scores and predict initial relation detection results in the form of 5-tuples (Sec.~\ref{sec:approach:scene_graph}).
These groundings are further refined using WSOD techniques \cite{Tang_2017_CVPR}  (Sec.~\ref{sec:approach:iterative}).
Finally, we capture visuo-linguistic common sense to further rule out unlikely relation tuples  (Fig.~\ref{fig:overview} middle-bottom, Sec.~\ref{sec:approach:sequential_context}). We use an RNN to model the fluency of scene graph tuples, enforcing that subject/object regions should be followed by their labels, 
and subject/object should be followed by object/predicate. 
This module reassigns labels and reranks 5-tuples to improve the relation detection: if an uncommon tuple is fed to the model, it will be assigned a low score.  


\subsection{Modeling \textbf{\textsc{Phrasal}} context}
\label{sec:approach:phrasal_context}

We first determine how to represent the text entities to be matched in the image. A naive solution would be to use the word embeddings, but this method ignores the context captured in phrases.
We advocate the use of the hints in the phrasal structure, namely mentions of related adjectives and objects.
As shown in Fig.~\ref{fig:overview} top-left, ``wearing sunglasses,'' ``sitting on the sofa'' and ``eating a banana'' provide context for the same ``girl'' and make her distinguishable from other potential instances of ``girl''. 
We infer the contextualized entity word features via the phrasal context and apply them in Sec.~\ref{sec:approach:grounding} to localize visual objects.




We have summarized all notations in Tab.~\ref{tab:notations} to facilitate reading the following text.
The linguistic structure (Fig.~\ref{fig:overview} top-left) parsed from a caption is represented using a text graph $G^L=(E, R)$.
$E=[e_1 \cdots e_{n_e}]^T$ denotes the $n_e$ text graph entities where each $e_i \in \{1 \dots c_e\}$ represents an entity class ID ($c_e$ classes in total,
which are defined by \cite{Zareian_2020_CVPR} or \cite{Xu_2017_CVPR} in our experiments;
in Fig.~\ref{fig:overview} top-left, $E=[\text{``glasses'', ``girl'', ``banana'', ``sofa''}]^T$).
$R=[(r_1, s_1, o_1) \cdots (r_{n_r}, s_{n_r}, o_{n_r})]^T$ describes the $n_r$ relations. For the $i$-th relation: $r_i\in \{1 \dots c_r\}$ is the relation class ID;  $s_i, o_i \in \{1 \dots n_e\}$ are entity indices:
$e_{s_i}$ denotes the subject entity and $e_{o_i}$ the object entity; in Fig.~\ref{fig:overview} top-left, $R=\text{\{(``wear'', 2, 1), (``eat'', 2, 3), (``sit'', 2, 4))\}}$.
Given the GloVe embedding~\cite{pennington-etal-2014-glove} of the entity and relation classes $W_{ent} \in \mathbb{R}^{c_e \times d}$, $W_{rel} \in \mathbb{R}^{c_r \times d}$, and the one-hot representation of entities and relations $Y_{ent} \in \mathbb{R}^{n_e\times c_e}$, $Y_{rel} \in \mathbb{R}^{n_r\times c_r}$ (each row is a $c_e$ or $c_r$-dim one-hot vector, and there are $n_e$ and $n_r$ rows, respectively), the initial entity and relation word embeddings can be represented as $H_{ent}^{(0)}=Y_{ent} W_{ent} \in \mathbb{R}^{n_e \times d}$ and $H_{rel}^{(0)}=Y_{rel} W_{rel} \in \mathbb{R}^{n_r \times d}$.

Now we compute phrasal contextualized entity embeddings $\psi(E;G^L) \in \mathbb{R}^{n_e \times d}$.
Alg.~\ref{alg:message_passing_neural_network} shows the process, and can be stacked several times. 
We update relation edge embeddings, then aggregate the relation features into the connected entity nodes, 
using linear layers $\phi^r$ and $\phi^{\alpha}$ applied on the concatenation of inputs. 
We use $\psi(E; G^L)=H_{ent}^{(t)}, (t>1)$ in the next section, to localize visual entities.

\newcommand\codecomment[1]{\footnotesize\textcolor{blue}{#1}}
\begin{algorithm}
\caption{Message passing to utilize phrasal context. We use TF-GraphNets~\cite{battaglia2018relational} to implement.}
\label{alg:message_passing_neural_network}
\SetKw{Let}{let}
\SetKwInOut{Input}{Input}
\SetKwInOut{Output}{Output}
\Input{Text graph $G^L=(E, R)$ \\
       Initial entity features $H_{ent}^{(t)}=[\bm{e}_1, \dots, \bm{e}_{n_e}]^T$ \\
       Initial relation features $H_{rel}^{(t)}=[\bm{r}_1, \dots, \bm{r}_{n_r}]^T$}
\Output{Updated $H_{ent}^{(t+1)}$, $H_{rel}^{(t+1)}$}
\For{$i \gets 1$ to $n_r$} {
    $\bm{r}_i' \gets \phi^r(\bm{r}_i, \bm{e}_{s_i}, \bm{e}_{o_i})$ 
    \codecomment{~// Update edge, $\bm{r}_i' \in \mathbb{R}^{d\times 1}$}\\
    $\alpha_i \gets \phi^{\alpha}(\bm{r}_i, \bm{e}_{s_i}, \bm{e}_{o_i})$
    \codecomment{~// Update edge weight, $\alpha_i \in \mathbb{R}^1$}
}
\For{$i \gets 1$ to $n_e$} {
    $\bm{e}_i'\!\gets\!\!\!\!\!\sum\limits_{\substack{j=1:n_r,\\o_j=i}} \!\!\!\!\big\{\frac{\exp(\alpha_j)}{\sum\limits_{\substack{k=1:n_r,\\o_k=i}}\exp(\alpha_k)} \big\}
    \bm{r}_{j}'$
    \codecomment{~//Aggregate, $\bm{e}_i' \in \mathbb{R}^{d\times 1}$}
}
\Return $H_{ent}^{(t+1)}=[\bm{e}_1' \cdots \bm{e}_{n_e}']^T, H_{rel}^{(t+1)}=[\bm{r}_1' \cdots \bm{r}_{n_r}']^T$
\end{algorithm}


\subsection{Associating text entities with visual boxes}
\label{sec:approach:grounding}

After getting the contextualized entity embeddings $\psi(E; G^L) \in \mathbb{R}^{n_e \times d}$, we seek their associated visual regions $\bm{g}^{(0)} \in \mathbb{R}^{n_e \times 1}$ (i.e., grounding vector), where each $\bm{g}_i^{(0)}$ ranges in $\{1 \cdots n_v\}$ and $v_{\bm{g}_i^{(0)}}$ denotes the visual box best describing the text entity $e_i$.
We obtain $\bm{g}$ using an attention mechanism. By optimizing the image-level prediction, we expect the model to learn to focus on the most informative and distinguishable regions, which can often be used as instance references for training object detectors.

We first project 
$V_{feat} \in \mathbb{R}^{n_v \times d_{cnn}}$ to the $d$-dim visual-language space,
resulting in attention and classification heads $H_{att}, H_{cls} \in \mathbb{R}^{n_v \times d}$.
Then, we 
compute 
$D_{dot}\in \mathbb{R}^{n_e \times n_v}$, in which $D_{dot}[i,j]$ measures the compatibility between text entity $e_i$ and visual region $v_j$.
We softmax-normalize $D_{dot}$ to get the attention matrix $A^{(0)}\in \mathbb{R}^{n_e \times n_v}$,
and obtain $\bm{g}^{(0)}$ by selecting the max-valued entry.

\vspace{-0.5cm}
\begin{align}
    &H_{att} = V_{feat} W_{att}, ~ H_{cls} = V_{feat} W_{cls} \notag \\
    &D_{dot} = \psi(E;G^L) H_{att}^T, ~ A^{(0)}[i,j]=\frac{\exp(D_{dot}[i,j])}{\sum_{k=1}^{n_v} \exp(D_{dot}[i,k])} \notag \\
    &\bm{g}_i^{(0)} = \argmax_{j \in \{1 \cdots n_v\}} A^{(0)}[i,j]  \label{eq:grounding_by_attention}
\end{align}


We use image-level entity labels $Y_{ent} \in \mathbb{R}^{n_e \times c_e}$ 
as supervision to learn proper attention scores. 
We first aggregate the image-level weighted visual features $F=[\bm{f}_1 \cdots \bm{f}_{n_e}]^T\in \mathbb{R}^{n_e \times d}$, 
where $\bm{f}_i$ denotes the image-level feature encoded with proper attention to highlight text entity $e_i$. For example, given $e_i=$~``glasses'' in Fig.~\ref{fig:overview}, the model needs to shift attention to the glasses visual region by adjusting the $i$-th row of $A^{(0)}$.
The final image-level entity classification score is given by $P_{cls}\in \mathbb{R}^{n_e \times c_e}$, and
the grounding module is trained using cross-entropy.

\vspace{-0.5cm}
\begin{align} 
    \begin{split} \label{eq:grounding}
        &F=A^{(0)} H_{cls}, ~ F'=FW_{ent}^T \\
        &P_{cls}[i,j] = \frac{\exp (F'[i,j])}{\sum_{k=1}^{c_e} \exp(F'[i,k])}
    \end{split} \\
    & L_{grd} = - \sum_{i=1}^{n_e} \sum_{j=1}^{c_e}  Y_{ent}[i,j]  \log P_{cls}[i,j] \label{eq:grounding_loss}
\end{align}

\subsection{Initial scene graph generation}
\label{sec:approach:scene_graph}

Thus far, the text entity embeddings $H_{ent}^{(0)}$ played a role in the grounding procedure, and so did the one-hot encoded label $Y_{ent}$ extracted from the caption. Next, 
the model learns to predict the entities and relations without help from captions, which will not be available at inference time.

To this end, given entities $E=[e_1 \cdots e_{n_e}]^T$, relations $R=[(r_1, s_1, o_1) \cdots (r_{n_r}, s_{n_r}, o_{n_r})]^T$, and grounded boxes $[v_{\bm{g}_1^{(0)}} \cdots v_{\bm{g}_{n_e}^{(0)}}]^T$, we first parse the \emph{target} instance labels. 
We extract $Y_{det}^{(0)} \in \mathbb{R}^{n_v \times c_e}$ and $Y_{relsub}, Y_{relobj} \in \mathbb{R}^{n_v \times c_r}$ using Eq.~\ref{eq:pseudo_labels}, in which all non-mentioned matrix entries are set to $0$.
$Y_{det}^{(0)}[i,j]=1$ means visual region $v_i$ involves the $j$-th entity class.
$Y_{relsub}[i,j]=1$ denotes the potential subject visual region $v_i$ (e.g. a ``person'' region) may apply the $j$-th relation (e.g. ``ride'') to an unknown object.
$Y_{relobj}[i,j]=1$ denotes an unknown subject may apply the $j$-th relation to the potential object visual region $v_i$ (e.g. a ``horse'' region).
We add $_{rel}$ to highlight $Y_{relsub}, Y_{relobj}$ are relation instance-level labels, but are attached to the grounded subject and object visual boxes respectively. 

\vspace{-0.5cm}
\begin{align} \label{eq:pseudo_labels}
    &Y_{det}^{(0)}[i,j] = 1 ~ \text{if} ~ \exists k \in \{1 \cdots n_e\}, s.t. (\bm{g}^{(0)}_k = i, e_k = j) \notag \\
    &Y_{relsub}[i,j] = 1 ~ \text{if} ~ \exists k \in \{1 \cdots n_r\}, s.t. (\bm{g}^{(n_t)}_{s_k} = i, r_k = j) \notag \\
    &Y_{relobj}[i,j] = 1 ~ \text{if} ~ \exists k \in \{1 \cdots n_r\}, s.t. (\bm{g}^{(n_t)}_{o_k} = i, r_k = j)
\vspace{-0.5cm}
\end{align}

We next learn to predict the instance-level labels based on these targets, 
using entity detection head $H_{det}^{(0)} \in \mathbb{R}^{n_v \times d}$, and relation detection heads $H_{relsub}, H_{relobj} \in \mathbb{R}^{n_v \times d}$. Then, we matrix-multiply the three heads to the entity embedding $W_{ent} \in \mathbb{R}^{c_e \times d}$ and relation embedding $W_{rel}\in \mathbb{R}^{c_r \times d}$, and softmax-normalize, 
resulting in entity detection scores $P_{det}^{(0)}\in \mathbb{R}^{n_v\times c_e}$ and subject/object detection scores $P_{relsub}, P_{relobj}\in \mathbb{R}^{n_v\times c_r}$.
We use cross-entropy loss terms $L_{det}^{(0)}, L_{relsub}, L_{relobj}$ similar to Eq.~\ref{eq:grounding_loss} to approximate $P_{det}^{(0)} \sim Y_{det}^{(0)}$, $P_{relsub} \sim Y_{relsub}$, and $P_{relobj} \sim Y_{relobj}$.

\vspace{-0.3cm}
\begin{align} \label{eq:detection_scores}
    \begin{split}
        & X \in \{det, relsub, relobj\}, ~ W' \in \{W_{ent}, W_{rel}\} \\
        & H_X=V_{feat}W_X, F_X=H_X W'^T \\ 
        & P_X[i,j]=\frac{\exp(F_X[i,j])}{\sum_k\exp(F_X[i,k])}
    \end{split}
\end{align}

After training the aforementioned model, we can detect entities using $P_{det}^{(0)}\in \mathbb{R}^{n_v\times c_e}$ and detect relations using $P_{rel}\in \mathbb{R}^{n_v\times n_v\times c_r}$, where $P_{rel}[i,j,k]=\min(P_{relsub}[i,k], P_{relobj}[j, k])$.
Intuitively, 
we treat the relation as valid if it could be both implied from the subject and object visual regions. For example, if the model infers ``ride'' from the ``person'' region and estimates ``ride'' can also apply to object region ``horse'', it determines that ``ride'' is the proper predicate bridging the two regions. \cite{Zareian_2020_CVPR,Zhang_2017_ICCV}  proposed similar architectures to infer relation from a single region, \cite{Zareian_2020_CVPR} for optimizing runtime and \cite{Zhang_2017_ICCV} to avoid bad solutions. We use this idea because it is simple and effective, in combination with our stronger module in Sec.~\ref{sec:approach:sequential_context}.


\textbf{Test time post-processing.} 
Given $P_{det}^{(0)}$, and $P_{rel}$, we adopt the \textit{top-K predictions} (in experiments, $k$=$50,100$) denoted in Eq.~\ref{eq:initial_sggen} as the initial scene graph generation (SGGen) results. In Eq.~\ref{eq:initial_sggen}, the universal set $U=\{(v_{s_i^v}, v_{o_i^v}, s_i^e, p_i^r, o_i^e)\}_i$ denotes all possible 5-tuple combinations and $B$ is a subset of $U$ of size $k$. 
The goal is to seek the subset $B (B \subset U \text{ and } |B|=k)$ such that the sum of log probabilities is maximized. Within a specific $B$, $s^v, o^v \in \{1 \cdots n_v\}$ are the indices of 
proposal boxes to represent the subject and object regions, respectively; $s^e, o^e \in \{1 \dots c_e\}$ are subject and object entity class IDs; $p^r \in \{1 \dots c_r\}$ is the relation class ID. To implement Eq.~\ref{eq:initial_sggen} in practice, we use non-max suppression on $P_{det}^{(0)}$ to 
reduce the search space (ruling out unlikely classes and boxes).

\vspace{-0.5cm}
\begin{align} \label{eq:initial_sggen}
    SG_{init} &= \argmax_{B \subset U, |B|=k} \sum_{(s^v, o^v, s^e, p^r, o^e)\in B}
    \Big(\log P_{det}^{(0)}[s^v, s^e] \notag \\ 
        &+ \log P_{rel}[s^v, o^v, p^r] + \log P_{det}^{(0)}[o^v, o^e]
    \Big)
\end{align}



\subsection{\textsc{\textbf{Iterative}} detection scores estimation}
\label{sec:approach:iterative}

Careful readers may notice the superscript $^{(0)}$ in grounding vector $\bm{g}^{(0)}$, attention $A^{(0)}$, instance label $Y_{det}^{(0)}$, and instance prediction $P_{det}^{(0)}$. We use the superscript $^{(0)}$ to denote these are initial grounding results, which could be improved by the WSOD iterative refining technique proposed in \cite{Tang_2017_CVPR}. Suppose loss $L_{det}^{(t)}$ ($t\ge 0$) 
brings $P_{det}^{(t)}\in \mathbb{R}^{n_v \times c_e}$ close to $Y_{det}^{(t)}\in \mathbb{R}^{n_v\times c_e}$, where $Y_{det}^{(t)}$ is the caption-guided target label and $P_{det}^{(t)}$ is the prediction without help from captions.
We could then incorporate the entity information $E=[e_1 \cdots e_{n_e}]^T$ of the caption into $P_{det}^{(t)}$ 
to turn it into a stronger instance-level label $Y_{det}^{(t + 1)}$. 
The motivation is that the initial label $Y_{det}^{(0)}$ extracted from attention (Eq.~\ref{eq:grounding_by_attention}, \ref{eq:pseudo_labels}) will be easily influenced by the \textit{noise in captions}. Since the attention scores always sum to one,
some region will be assigned a higher score than others, regardless of whether the objects have consistent visual appearance. In an extreme case, mentioned but not visually present entities also have a matched proposal. Using $P_{det}^{(t)}$ is an indirect way to also consider the visual model's (Eq.~\ref{eq:detection_scores}) output, which encodes the
objects' consistent appearance.

To turn $P_{det}^{(t)}$ into $Y_{det}^{(t+1)}$, we first extract $A^{(t+1)} \in \mathbb{R}^{n_e \times n_v}$ (same shape as the attention matrix $A^{(0)}$). We simply select the columns (denoted as $[:, i]$) from $P_{det}^{(t)}$ according to $E$ to achieve $A^{(t+1)}$, and compute $\bm{g}^{(t+1)}$ and $Y_{det}^{(t+1)}$.

\vspace{-0.5cm}
\begin{align} \label{eq:oicr}
    &A^{(t+1)}=\Big[P_{det}^{(t)}[:,e_1] \cdots P_{det}^{(t)}[:,e_{n_e}]\Big]^T \notag \\
    &\bm{g}^{(t + 1)}=\argmax_{j\in\{1 \cdots n_v\}} A^{(t + 1)}[i,j] \\
    &Y_{det}^{(t+1)}[i,j] = 1 ~ \text{if} ~ \exists k \in \{1 \cdots n_e\}, s.t. (\bm{g}^{(t+1)}_k = i, e_k = j) \notag
\end{align}

We 
refine the model $n_t$ times, and 
in Eq.~\ref{eq:pseudo_labels}, we use $\bm{g}^{(n_t)}$ from the last iteration to compute $Y_{relsub}$ and $Y_{relobj}$. 


\subsection{Modeling \textsc{\textbf{Sequential}} context}
\label{sec:approach:sequential_context}

We observed the model sometimes generates triplets that violate common sense, e.g., plate-on-pizza in Fig.~\ref{fig:sequential_qualitative} top, because the aforementioned test time post-processing (Eq.~\ref{eq:initial_sggen}) considers predictions from $P_{det}$ and $P_{rel}$ separately. When joined, the results may not form a meaningful triplet. To solve the problem, we propose a vision-language module to consider sequential patterns summarized from the dataset (Fig.~\ref{fig:overview} middle-bottom).
The idea is inspired by \cite{lu2016visual}, but different because: (1) we encode the language and vision priors within the same multi-modal RNN while \cite{lu2016visual} models vision and language separately, and (2) our label generation captures a language N-gram such that the later generated object and predicate will not contradict the subject.

Specifically, we gather the grounded tuples $D_{gt}=\{(v_{\bm{g}_{s_i}}, v_{\bm{g}_{o_i}}, e_{s_i}, r_i, e_{o_i})\}_{i=1}^{n_r}$ within each training example to learn the sequential patterns. Compared to the SGGen 5-tuple (Eq.~\ref{eq:initial_sggen}), the $e_{s_i}, r_i, e_{o_i}$ here are from the ground-truth $(E, R)$ and are always correct (e.g., no ``cake-eat-person''). Since the module receives high-quality supervision from captions, it will assign low scores or adjust the prediction (Eq.~\ref{eq:initial_sggen}) for imprecise 5-tuples at test time, using its estimate of what proper 5-tuples look like.

Fig.~\ref{fig:overview} middle-bottom shows the idea. 
We use an RNN (LSTM in our implementation) to consume both word embeddings and visual features of the subject and object. The training outputs are subject prediction $P_{cssub}\in \mathbb{R}^{n_r \times c_e}$ ($_{cs}$ for \emph{c}ommon \emph{s}ense), object prediction $P_{csobj}\in \mathbb{R}^{n_r\times c_e}$, and predicate prediction $P_{cspred}\in \mathbb{R}^{n_r\times c_r}$. We now explain how to generate their $i$-th row (to match true $e_{s_i}$-$r_i$-$e_{o_i}$).

First, we feed into the RNN a dummy \textit{/start/} embedding and the grounded subject visual feature $\bm{v}_{\bm{g}_{s_i}}$. The subject prediction $P_{cssub}[i,:]$ is achieved by a linear layer projection (from RNN output to $d$-dim) and matrix multiplication (using $W_{ent}\in\mathbb{R}^{c_e\times d}$).
We predict the object $P_{csobj}[i,:]$ similarly, but using the grounded object visual feature $\bm{v}_{\bm{g}_{o_i}}$ concatenated with the subject word embedding $e_{s_i}$ as inputs. If we do not consider the visual input, this step is akin to learning a subject-object 2-gram language model.
Next, the RNN predicts predicate label $P_{cspred}[i, :]$ (using $W_{rel}\in \mathbb{R}^{c_r\times d}$ instead of $W_{ent}$), using object word embedding $e_{o_i}$ and a dummy visual feature $\emptyset$ as inputs.

To learn $P_{cssub}, P_{csobj}, P_{cspred}$, we extract labels $Y_{cssub}, Y_{csobj}, Y_{cspred}$ 
(Eq.~\ref{eq:common_sense_labels}) and use cross-entropy losses $L_{cssub} (P_{cssub} \sim Y_{cssub})$, $L_{csobj} (P_{csobj} \sim Y_{csobj})$, $L_{cspred} (P_{cspred} \sim Y_{cspred})$ to optimize the RNN model. 

\vspace{-0.5cm}
\begin{align} \label{eq:common_sense_labels}
    &Y_{cssub}=\Big[Y_{ent}[e_{s_i},:]^T \cdots Y_{ent}[e_{s_{n_r}},:]^T \Big]^T\\
    &Y_{csobj}=\Big[Y_{ent}[e_{o_i},:]^T \cdots Y_{ent}[e_{o_{n_r}},:]^T \Big]^T, ~ Y_{cspred}=Y_{rel} \notag
\end{align}

At test time, we feed to the RNN the visual features from $SG_{init}$ (Eq.~\ref{eq:initial_sggen}) and the \textit{/start/} embedding. We let the RNN re-label the subject-object-predicate using beam search.
The final score for each re-labeled 5-tuple is the sum of log probabilities of generating subject, object, and predicate. 
We generate the object before the predicate because objects are usually more distinguishable than predicates, so this order simplifies inference, allowing the use of a smaller beam size.
We re-rank the beam search results 
using the final scores and keep the top ones to compute the Recall@$k$ to evaluate (examples in Fig.~\ref{fig:sequential_qualitative}, Fig.~\ref{fig:basic_vs_final}).

\textbf{Our final model} is trained using the following multi-task loss, where $\beta$ is set to $0.5$ since at the core of the task is the grounding of visual objects.

\vspace{-0.5cm}
\begin{align} \label{eq:final_loss}
    L = & L_{grd} + \beta \big( \sum_{t=0}^{n_t}L_{det}^{(t)} + L_{relsub} + L_{relobj} \notag \\
        &+ L_{cssub} + L_{csobj} + L_{cspred}\big)
\end{align}

\section{Experiments}
\label{sec:results}

\textbf{Datasets.}
We use the Visual Genome (VG)~\cite{krishna2017visual} and Common Objects in Context (COCO)~\cite{lin2014microsoft} datasets, which both provide captions describing the visual contents.
VG involves 108,077 images and 5.4 million region descriptions. The associated annotations of 3.8 million object instances and 2.3 million relationships enable us to evaluate the scene graph generation performance.
To fairly compare to the counterpart weakly-supervised scene graph generation methods \cite{Zhang_2017_ICCV,Zareian_2020_CVPR}, we adopt the VG split used in \underline{Zareian \etal}~\cite{Zareian_2020_CVPR}: keeping the most frequent $c_e=200$ entity classes and $c_r=100$ predicate classes, resulting in 99,646 images with \textit{subject-predicate-object} annotations. We use the same 73,791/25,855 train/test split\footnote{We follow \cite{Zareian_2020_CVPR}, but \cite{Zhang_2017_ICCV} reports 73,801/25,857 train/test split}.
We also adopt the split in \underline{Xu \etal}~\cite{Xu_2017_CVPR}, more commonly used by fully-supervised methods. It contains 75,651/32,422 train/test images and keeps $c_e=150$ entity and $c_r=50$ predicate classes.
Both VG splits are preprocessed by \cite{Zareian_2020_CVPR}.

For COCO data, we use the 2017 training split (118,287 images). We rule out the duplicated images in the VG test set, resulting in 106,401 images for Zareian \etal's split and 102,786 images for Xu \etal's.

\textbf{Learning tasks.}
The linguistic structure supervision for training is from the following three sources:

\begin{itemize}[noitemsep,nolistsep,leftmargin=4.3mm]
    \item \underline{VG-GT-Graph} imagines an ideal scenario (an upper bound with the noise in captions and parsers' impacts isolated) where we have the ground-truth text graph annotations instead of a set of image-level \textit{subject-predicate-object} triplets, for training on VG. To get these ground-truth graphs, we check the visual regions associated with the entities (subjects and objects) and connect entities if their regions have IoU greater than 0.5. We do \emph{not} use box annotations to improve detection results.
    \item \underline{VG-Cap-Graph} utilizes the VG \textit{region} descriptions. We use \cite{schuster-etal-2015-generating} to extract text graphs from these descriptions, but we ignore the region coordinates and treat the graphs as image-level annotations.
    \item \underline{COCO-Cap-Graph} uses captions from COCO and applies the same parsing technique as VG-Cap-Graph. The difference is that these captions are image-level, and describe the objects and relations as a whole.
\end{itemize}

\textbf{Metrics.} 
We measure how accurately the models generate scene graphs, using the densely-annotated scene graphs in the VG test set. Following \cite{Xu_2017_CVPR}, a predicted triplet is considered correct if the three text labels are correct and the boxes for subject/object have $\ge 0.5$ IoU with ground-truth boxes. We then compute the Recall@50 and Recall@100 as the fraction of the ground-truth triplets that are successfully retrieved in the top-50 and top-100 predictions, respectively.

\textbf{Methods compared.}
We conduct ablation studies to verify the benefit of each component of our method.
\begin{itemize}[noitemsep,nolistsep,leftmargin=4.3mm]
    \item \textsc{\underline{Basic}} model refers to our Sec.~\ref{sec:approach:grounding}-\ref{sec:approach:scene_graph} without applying the phrasal contextualization. We set $\psi(E, G^L)=H_{ent}^{(0)}$.
    \item \textsc{\underline{+Phrasal}} context (Sec.~\ref{sec:approach:phrasal_context}) uses contextualized entity embeddings $\psi(E, G^L)$ instead of $H_{ent}^{(0)}$.
    \item \textsc{\underline{+Iterative}} (Sec.~\ref{sec:approach:iterative}) gradually improves the grounding vector $\bm{g}$. We iterate $n_t=3$ times by default.
    \item \textsc{\underline{+Sequential}} context (Sec.~\ref{sec:approach:sequential_context}) revises the prediction presented in Eq.~\ref{eq:initial_sggen}, using the RNN encoded with knowledge regarding sequential patterns.
\end{itemize}

We compare to weakly-supervised scene graph generation methods that published results on Zareian \etal's split: VtransE-MIL~\cite{zhang2017visual}, PPR-FCN-single \cite{Zhang_2017_ICCV}, PPR-FCN \cite{Zhang_2017_ICCV} 
and VSPNet \cite{Zareian_2020_CVPR}.
We also compare to fully-supervised methods on Xu \etal's split: Iterative Message Passing (IMP)~\cite{Xu_2017_CVPR}, Neural Motif Network (MotifNet)~\cite{Zellers_2018_CVPR}, Associative Embedding (Asso.Emb.)~\cite{Newell_2017_NeurIPS}, Multi-level Scene Description Network (MSDN)~\cite{Li_2017_ICCV}, Graph R-CNN~\cite{Yang_2018_ECCV}, and fully-supervised VSPNet~\cite{Zareian_2020_CVPR}.

\subsection{Results on GT-Graph setting}
\vspace{-0.2cm}


\begin{table}[t]
    \footnotesize
    \centering
    \setlength\tabcolsep{0pt} 
    
    \begin{tabularx}{1\linewidth}{c|c}
        \Xhline{2\arrayrulewidth}
        {\begin{tabularx}{0.5\linewidth}{c|*{2}{>{\centering\arraybackslash}X}}
            \multicolumn{3}{c}{Zareian \etal's split (weakly sup)} \\
        \Xhline{1\arrayrulewidth}
            Method & R@50 & R@100 \\
        \Xhline{1\arrayrulewidth}
                                                    &      &      \\
            VtranE-MIL~\cite{zhang2017visual}       & 0.71 & 0.90 \\
            PPR-FCN-single~\cite{Zhang_2017_ICCV}   & 1.08 & 1.63 \\
            PPR-FCN~\cite{Zhang_2017_ICCV}          & 1.52 & 1.90 \\
            VSPNet~\cite{Zareian_2020_CVPR}         & 3.10 & 3.50 \\
                                                    &      &      \\
        \Xhline{1\arrayrulewidth}
            \textsc{Basic}                          & 2.20 & 2.88 \\
            + \textsc{Phrasal}                      & 2.77 & 3.62 \\
            + \textsc{Iterative}                    & 3.26 & 4.15 \\
            + \textsc{Sequential}                   & 4.92 & 5.84 \\
        \end{tabularx}}
        &
        {\begin{tabularx}{0.5\linewidth}{c|*{2}{>{\centering\arraybackslash}X}}
            \multicolumn{3}{c}{Xu \etal's split (fully sup)} \\
        \Xhline{1\arrayrulewidth}
            Method & R@50 & R@100 \\
        \Xhline{1\arrayrulewidth}
            IMP~\cite{Xu_2017_CVPR}                 & 3.44  & 4.24  \\
            MotifNet~\cite{Zellers_2018_CVPR}       & 6.90  & 9.10  \\
            Asso.Emb.~\cite{Newell_2017_NeurIPS}    & 9.70  & 11.30 \\
            MSDN~\cite{Li_2017_ICCV}                & 10.72 & 14.22 \\
            Graph R-CNN~\cite{Yang_2018_ECCV}       & 11.40 & 13.70 \\
            VSPNet (Full)~\cite{Zareian_2020_CVPR}  & 12.60 & 14.20 \\
        \Xhline{1\arrayrulewidth}
            \textsc{Basic}                          & 3.82 & 4.96 \\
            + \textsc{Phrasal}                      & 4.04 & 5.21 \\
            + \textsc{Iterative}                    & 6.06 & 7.60 \\
            + \textsc{Sequential}                   & 7.30 & 8.73 \\
        \end{tabularx}} \\
        \Xhline{2\arrayrulewidth}
    \end{tabularx}
    \caption{SGGen recall (\%) under VG-GT-Graph setting. We compare our method to the state-of-the-art methods. High recall (R@50, R@100) is good.}
    \label{tab:gt_graph}
\end{table}

\begin{table}[t]
    \footnotesize
    \centering
    \setlength\tabcolsep{0pt} 
    \begin{tabularx}{1.05\linewidth}{c|*{2}{>{\centering\arraybackslash}X}|*{2}{>{\centering\arraybackslash}X}|*{2}{>{\centering\arraybackslash}X}|*{2}{>{\centering\arraybackslash}X}}
        \Xhline{2\arrayrulewidth}
            & \multicolumn{4}{c|}{VG-Cap-Graph} & \multicolumn{4}{c}{COCO-Cap-Graph} \\
        \Xhline{1\arrayrulewidth}
            \multirow{2}{*}{Eval split} & \multicolumn{2}{c|}{Zareian \etal's}
            & \multicolumn{2}{c|}{Xu \etal's}
            & \multicolumn{2}{c|}{Zareian \etal's}
            & \multicolumn{2}{c}{Xu \etal's} \\
            & R@50 & R@100
            & R@50 & R@100
            & R@50 & R@100
            & R@50 & R@100 \\
        \Xhline{1\arrayrulewidth}
            \textsc{Basic}          & 0.81 & 0.91 & 0.99 & 1.09 & 1.20 & 1.51 & 2.09 & 2.63 \\
            + \textsc{Phrasal}      & 0.90 & 1.04 & 1.39 & 1.69 & 1.17 & 1.47 & 1.65 & 2.16 \\
            + \textsc{Iterative}    & 1.11 & 1.32 & 1.79 & 2.22 & 1.41 & 1.75 & 2.41 & 3.02 \\
            + \textsc{Sequential}   & 1.83 & 1.94 & 3.85 & 4.04 & 1.95 & 2.23 & 3.28 & 3.69 \\
        \Xhline{2\arrayrulewidth}
    \end{tabularx}
    \caption{SGGen recall (\%) under Cap-Graph settings. High recall (R@50, R@100) is good.}
    \label{tab:cap_graph}
\end{table}

The GT-Graph setting allows our method to be fairly compared to the state-of-the-art methods because in this setting, the information ours and those methods receive is comparable (sets of triplets, in our case connected). Further, 
the word distribution is the same for training/testing, while the caption setting causes a train-test shift (described shortly).

In Tab.~\ref{tab:gt_graph} left, we show our results on Zareian \etal's VG split and baselines of weakly-supervised methods. Our \textsc{Basic} method already surpasses VtransE-MIL, PPR-FCN-single, and PPR-FCN. This may be due to the low quality of the EdgeBox proposals used in them.
Compared to VSPNet, which also uses Faster RCNN proposals, our \textsc{Basic} method is slightly worse, but our components greatly improve upon \textsc{Basic}, and our final model achieves 4.92, a 59\% improvement over VSPNet (using R@50).
\textsc{+Phrasal} context improves \textsc{Basic} by 26\% (2.77 v.s. 2.20), \textsc{+Iterative} improves \textsc{+Phrasal} by 18\% (3.26 v.s. 2.77), and \textsc{+Sequential} gains 51\% (4.92 v.s. 3.26).

In Tab.~\ref{tab:gt_graph} right, we compare to fully-supervised methods on Xu \etal's split. We observe our method is very competitive even though we only use image-level annotations. In terms of Recall@50, our final method (7.30) outperforms IMP (3.44) and MotifNet (6.90).
As for the relative improvement, \textsc{+Phrasal} context improves \textsc{Basic} by 6\% (4.04 v.s. 3.82), \textsc{+Iterative} gains 50\% (6.06 v.s. 4.04), and \textsc{+Sequential} gains 20\% (7.30 v.s. 6.06).

\subsection{Results on Cap-Graph setting}
\vspace{-0.2cm}

Our proposed Cap-Graph setting is an under-explored and challenging one, as the learned SGGen model depends on the captions' exhaustiveness and the parser's quality, but it allows learning from less expensive 
image-text data.

In Tab.~\ref{tab:cap_graph}, we show the SGGen performance of models learned from VG region captions (VG-Cap-Graph) and COCO image captions (COCO-Cap-Graph). We see the same trend as in GT-Graph setting: our components (\textsc{+Phrasal}, \textsc{+Iterative}, and \textsc{+Sequential}) have positive effects. 
Further, our final models learned from both VG-Cap-Graph (R@50 1.83) and COCO-Cap-Graph (1.95) are better than all weakly-supervised methods except VSPNet (in Tab.~\ref{tab:gt_graph} left). Our models learned from captions are even comparable (VG-Cap-Graph 3.85, COCO-Cap-Graph 3.28) to the fully-supervised IMP (R@50 3.44).

\begin{figure}[t]
    \centering
    \includegraphics[width=1.0\linewidth]{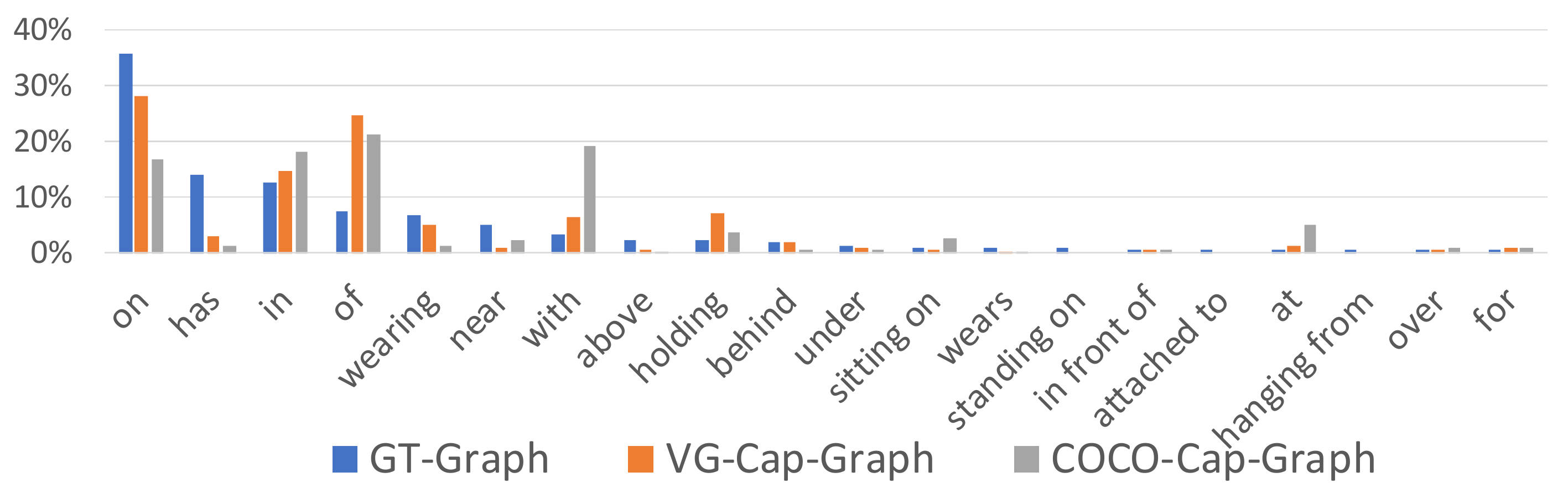}  
    \caption{Relation frequencies in the three settings.}
    \label{fig:relation_stats}
\end{figure}

Fig.~\ref{fig:relation_stats} shows the relation frequencies in our settings. We observe that some relations (``has,'' ``near'') rarely appear in text descriptions but are often annotated in ground-truth scene graphs. Meanwhile, there are frequently mentioned prepositions in captions (``of,'' ``with,'' ``at'') which are rarely denoted as relations. These train/test discrepancies (train on captions, test on triplets) explain our methods' relative performance in
Tab.~\ref{tab:cap_graph}, where \textsc{+Phrasal} helps under VG-Gap-Graph (more similar to GT-Graph) but hurts slightly under COCO-Cap-Graph (less similar to GT-Graph).

\subsection{Qualitative examples}
\label{sec:qualitative_examples}
\vspace{-0.2cm}

\begin{figure}[t]
    \centering
    \includegraphics[width=1.0\linewidth]{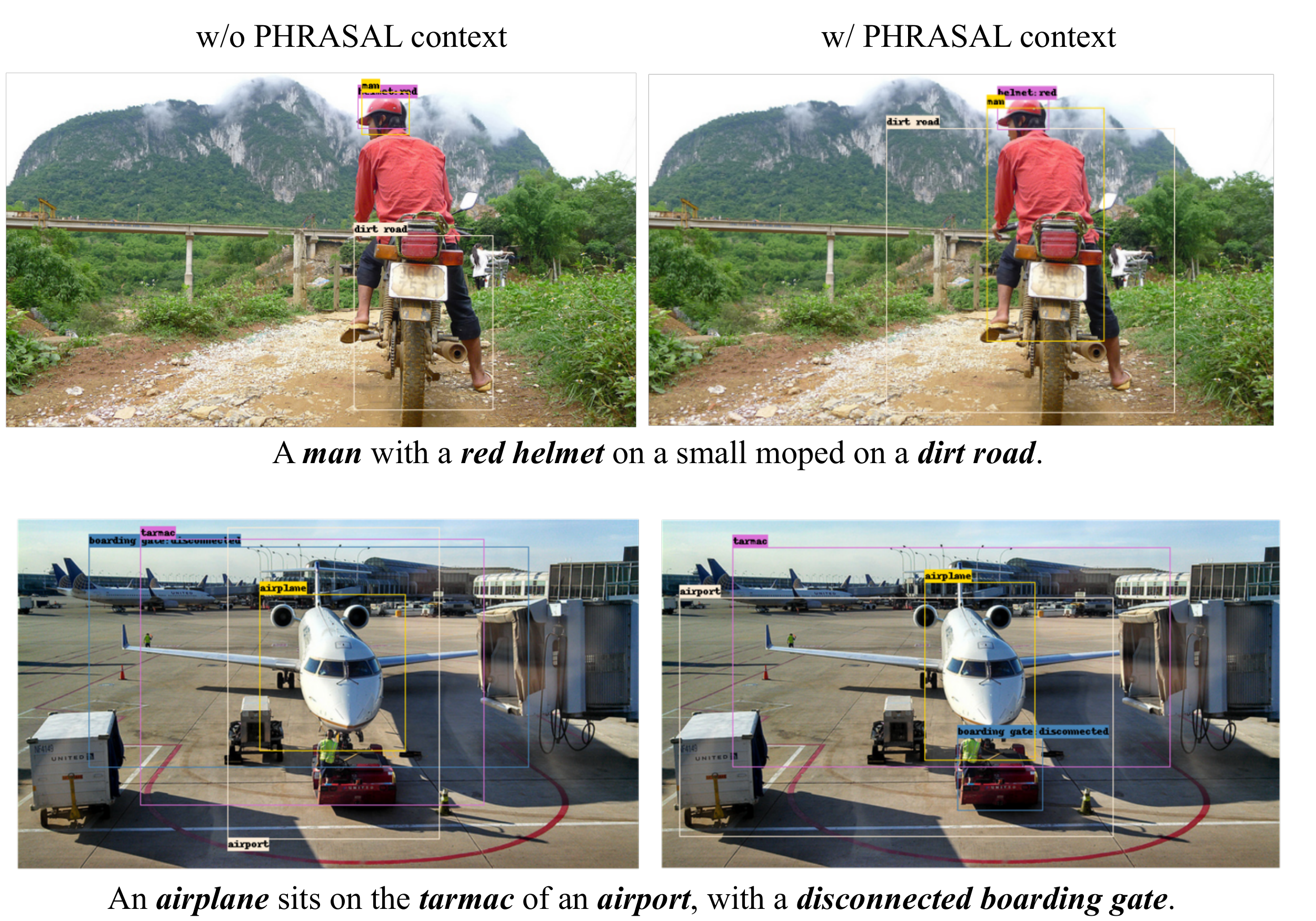}  
    \caption{Importance of \textsc{phrasal} context; best seen with zoom.}
    \label{fig:phrasal_qualitative}
\end{figure}

Fig.~\ref{fig:phrasal_qualitative} compares using and not using \textsc{Phrasal} context. Without out \textsc{Phrasal} module (left), the grounding procedure gets stuck on the same distinguishable local region (top-left: head of man) or erroneously attends to the whole image (bottom-left: boarding gate). When using the \textsc{Phrasal} module (right), our model is better at localizing visual objects. It knows there should be a complete person in the scene (top-right) and the boarding gate is a concept related to the plane (bottom-right). 

\begin{figure}[t]
    \centering
    \includegraphics[width=0.95\linewidth]{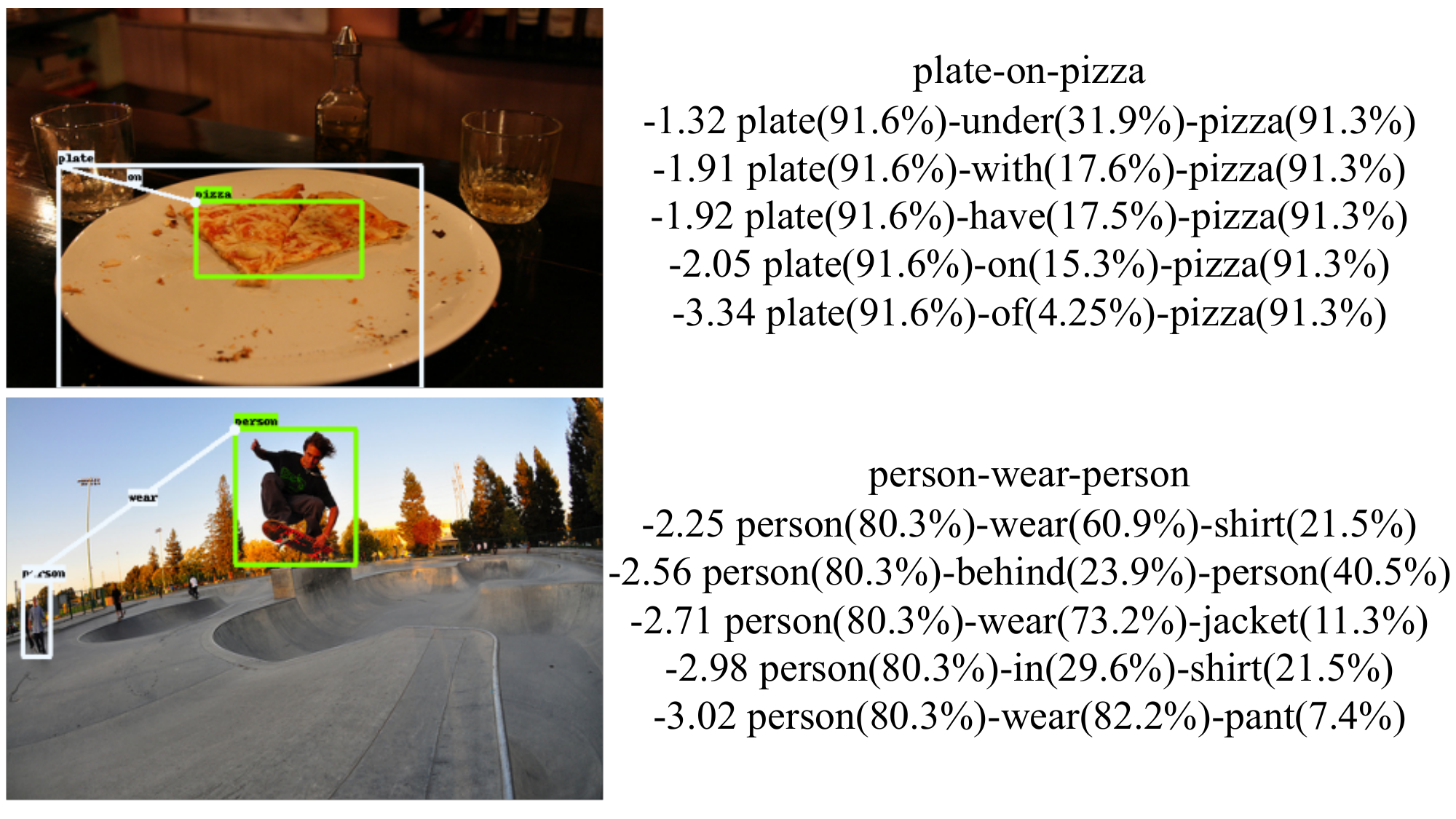}  
    \vspace{-0.1cm}
    \caption{Importance of \textsc{Sequential} context.}
    \label{fig:sequential_qualitative}
\end{figure}

\begin{figure}[t]
    \centering
    \includegraphics[width=0.95\linewidth]{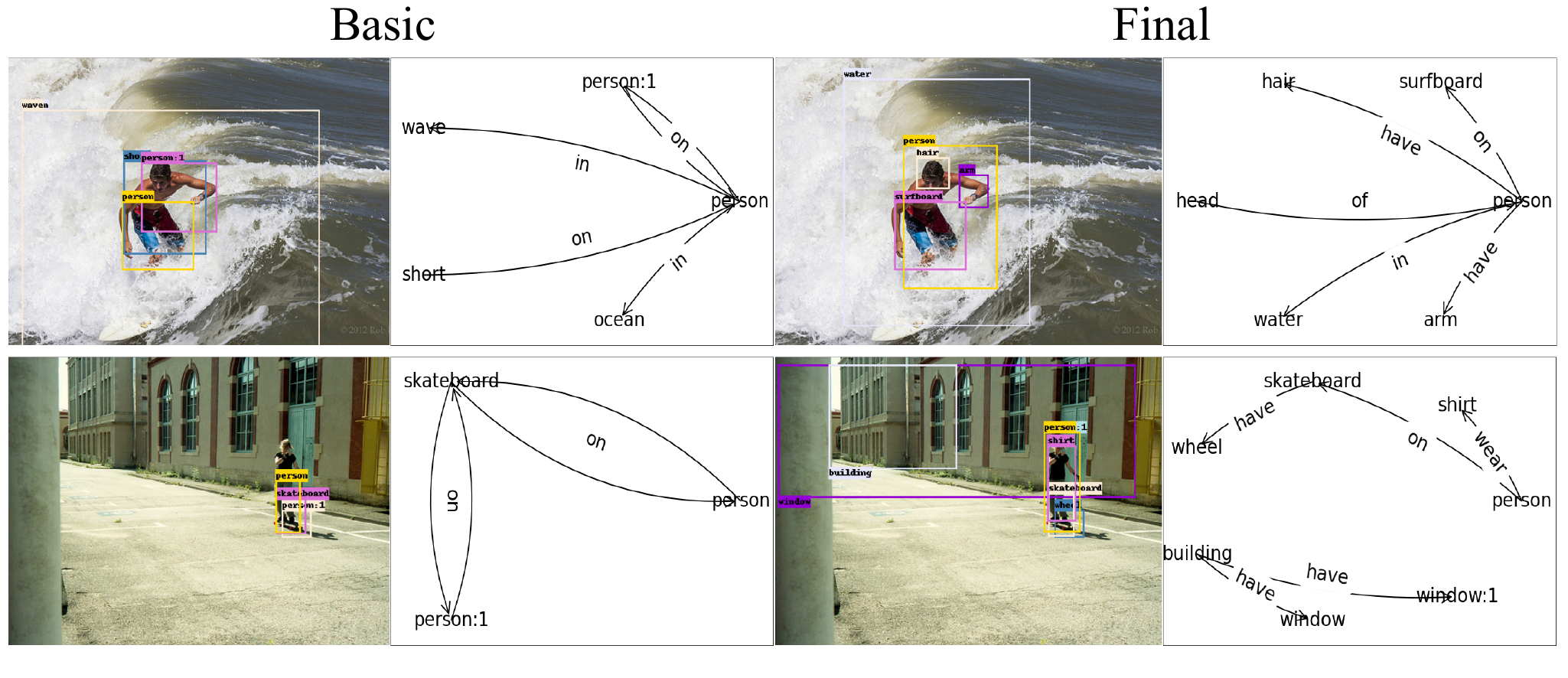}  
    \vspace{-0.1cm}
    \caption{\textsc{Basic} v.s. our final model; best viewed with zoom.}
    \label{fig:basic_vs_final}
\end{figure}

Fig.~\ref{fig:sequential_qualitative} shows how the learned sequential patterns help correct imprecise predictions. For the corrections (beam size=5), we show the log-probability of the 5-tuple and individual probabilities. 
Given that \emph{plate} cannot be put \emph{on pizza}, our model corrects it to \textit{plate-under-pizza}. In the bottom example, our model corrects \textit{person-wear-person} to \textit{person-wear-shirt} and \textit{person-behind-person}.
In Fig.~\ref{fig:basic_vs_final}, we compare our \textsc{Basic} and final methods.


\subsection{Implementation details}
\vspace{-0.2cm}


We pre-extract text graphs using \cite{Wu_2019_CVPR}'s implementation of \cite{schuster-etal-2015-generating}.
We use the same proposals ($n_v=20$ per image) and features ($d_{cnn}=1536$) as \cite{Zareian_2020_CVPR}, extracted using Faster-RCNN ~\cite{ren2015faster} (InceptionResnet backbone~\cite{szegedy2017inception}) pre-trained on OpenImage ~\cite{kuznetsova2020open}. 
During training, we use GraphNets~\cite{battaglia2018relational} to encode phrasal context. $W_{ent}$, $W_{rel}$ are $d=300$ frozen GloVe embeddings~\cite{pennington-etal-2014-glove}. 
To train our model, we use a batch size of 32, learning rate 0.00001, the Adam optimizer~\cite{kingma2014adam}, and Tensorflow distributed training~\cite{Abadi:2016:TSL:3026877.3026899}. 
We use weight decay of 1e-6 and the random normal initializer (mean=0.0, stdev=0.01) for all fully-connected layers. We use LSTM cell, 100 hidden units, and dropout 0.2, for the \textsc{Sequential} module.
For the non-max-suppression of Eq.~\ref{eq:initial_sggen}, we use score threshold 0.01, IoU threshold 0.4, and limit the maximum instances per entity class to 4.
We set beam size to 5 for the \textsc{Sequential} module post-processing.

\vspace{-0.2cm}
\section{Conclusion}
\label{sec:conclusion}
\vspace{-0.2cm}

We introduced a method that leverages caption supervision for scene graph generation. Captions are noisy, but can be obtained ``for free,'' and allow us to understand high-order relations between entities and triplets.
To leverage caption supervision, we proposed to capture commonsense relations (phrasal and sequential context) and iteratively refine detection scores. In the future, we will explicitly handle distribution shifts between captions and text graphs.

\noindent \textbf{Acknowledgements:} This work was partly supported by National Science Foundation Grant No. 1718262 and a Univ. of Pittsburgh Computer Science fellowship. We thank the reviewers and AC for their encouragement.
\clearpage

{\small
\bibliographystyle{ieee_fullname}
\bibliography{egbib}
}

\end{document}